\documentclass{ceurart}
\usepackage{amsmath,amssymb,amsthm}

\usepackage{listings}
\usepackage{comment}
\usepackage{subcaption}
\usepackage{graphicx}
\makeatletter\def\Hy@Warning#1{}\makeatother
\begin{document}

\copyrightyear{2024}
\copyrightclause{Copyright for this paper by its authors.
  }

\conference{Math UI 2024, Montreal Canada, August 9 2024}

\title{\mbox{A First Look at} \mbox{Chebyshev-Sobolev Series for Digital Ink}}

\author[1]{Deepak Singh Kalhan}[%
orcid=0009-0007-7585-2901,
email=dsinghkalhan@uwaterloo.ca,
url=www.linkedin.com/in/dsinghkalhan/,
]

\author[1]{Stephen M. Watt}[%
orcid=0000-0001-8303-4983,
email=smwatt@uwaterloo.ca,
url=https://cs.uwaterloo.ca/~smwatt/,
]
\address[1]{Cheriton School of Computer Science,
  University of Waterloo, Waterloo Canada, N2L 3G1}

\begin{abstract}
Considering digital ink as plane curves provides a valuable framework for various applications, including signature verification, note-taking, and mathematical handwriting recognition. 
These plane curves can be obtained as parameterized pairs of approximating truncated series $\big ( x(s), y(s) \big )$ determined by sampled points.
Earlier work has found that representing these truncated series (polynomials) in a Legendre or Legendre-Sobolev basis has a number of desirable properties.  These include 
compact data representation, 
meaningful clustering of like symbols in the vector space of polynomial coefficients,
linear separability of classes in this space,
and 
highly efficient calculation of variation between curves.
In this work, we take a first step at examining the use of Chebyshev-Sobolev series for symbol recognition.   The early indication is that this representation may be superior to Legendre-Sobolev representation for some purposes.
\end{abstract}


\maketitle
\section{Introduction}
\label{sec:intro}
Digital ink is an important tool in modern computing with applications ranging from note-taking, signature verification to handwriting and mathematical expression recognition. In order to accurately use digital ink data, methods like orthogonal series representations can be employed.
The use of orthogonal polynomials such as Legendre, Laguerre, Chebyshev or other polynomials provides a powerful framework for approximating functions. 
The Legendre-Sobolev series has gained attention for its improved recognition rates \cite{mazalov2012improving}. In this paper, we extend the exploration of orthogonal series representation by focusing on Chebyshev-Sobolev series. 

Viewing digital ink as parametric plane curves is a sharp departure from previous work which used either a pixel-based approach or point-sequence approach.  Both of these earlier approaches are resolution dependent, one in space and the other in time.   This leads to a host of challenges.  For example, older data is ``resampled'' (\textit{i.e.} interpolated, then discretized) to use these methods with data from different sources.  Another issue is in matching corresponding points in comparing curves, leading to challenging optimization of alignment-based ``dynamic time warping''.
Working with parametric curves as algebraic objects avoids all these problems.

Building on the results using Legendre-Sobolev series, this paper explores the potential of the Chebyshev-Sobolev series for digital ink representation. Chebyshev polynomials are known for stability \cite{wu2015new} and improved accuracy for functional approximation \cite{lanczos1952solution}. The Sobolev space improves shape matching by including the derivatives in the expression.
Therefore, we suspect Chebyshev-Sobolev series may be suitable for handwritten text and symbol recognition. 

This paper is organized as follows:
Section\ref{sec:background} discusses the background methods to use an orthogonal polynomial basis for digital ink representation. In Section \ref{sec:chebyshev-sobolev-ip}, we define a ``Chebyshev-Sobolev'' inner product and give some of its properties.
Section~\ref{sec:algorithm} gives an algorithm to classify a sample curve against a database of many reference curves.  This uses the same idea as earlier work based on Legendre-Sobolev polynomials.
Section~\ref{sec:experiments} describes some initial experiments examining the use of digital ink in a Chebyshev-Sobolev basis representation.
Finally, we give some brief conclusions in Section~\ref{sec:conclusions}.
 
\section{Background}
\label{sec:background}
Digital ink is typically generated by sampling points from pen traces which results in a series of points, each giving $x$ and $y$ coordinates. 
The time interval and resolution depend on the hardware so devices with different hardware will result in different series of points from the same input trace.  As technology evolves, various \textit{ad hoc} treatments have been developed, such as size normalization and resampling to deal with these differences.
However, these methods introduce their own problems.
Representing handwritten symbols in the space of coefficients of an orthogonal basis approximation is a useful alternative method \cite{watt2012polynomial, golubitsky2008online, char2007representing, golubitsky2010distance} and has proven to be robust against changes in hardware \cite{hu2013identifying}. Additionally, this method has high recognition rates with small training sets \cite{watt2012polynomial}. 

\subsection{Parameterized Curves in an Orthogonal Basis}
In our work, we have considered an ink trace as a segment of plane curve $x(s)$ and $y(s)$, parameterized by euclidean arc length $s$, where
\begin{align}
    ds^2 = dx^2 + dy^2.
\end{align}
If we let $\{B_i(s)\}_{i=0,....d}$ be a graded polynomial basis for approximating ($x(s)$, $y(s)$), a trace can be represented as 
\begin{align}
    &x(s) \approx \sum_{i=0}^{d} x_i B_i(s)  \\
    &y(s) \approx \sum_{i=0}^{d} y_i B_i(s)
\end{align}
where, $x_i$ and $y_i$ are the coefficients and $d$ is the degree of the truncated series. 
To allow efficient calculation of coefficients, the functions $B_i(s)$ can be chosen to be polynomials that are orthogonal with respect to a functional inner product.
The approximated curve can be made close to the original trace by choosing higher value of $d$. 

Given a functional inner product, a graded basis of orthogonal polynomials may be obtained using Gram-Schmidt orthogonalization of the set of monomials $\{s^i\}_{i \in [0..d]}$.
Then a digital ink trace can be represented in this basis as
the vector of coefficients $(x_0,....,x_d,y_0,.....,y_d)$. 

We can now illustrate the benefit or representing curves in an orthogonal basis.
 The usual way of comparing two traces given as series of sampled points 
 is to find a correspondence between points of the traces to minimize the sum of distances between corresponding points.
 Suppose two traces are given as point sequences $(X_0, Y_0), \ldots, (X_m, Y_m)$ and $(U_0,V_0), \ldots, (U_n, V_n)$.  (Note that here we use subscripts to indicate a position in the sequence, not a coefficient.)
 Without loss of generality, assume $m \ge n$.
 Then we measure the squared distance between the two traces as
 \begin{equation}
  \min_{\phi: [0..m] \rightarrow [0..n]}    \sum_{i=0}^m (X_i - U_{\phi (i)})^2 + (Y_i-V_{\phi (i)})^2
 \end{equation}
 over all choices of non-decreasing $\phi$ with $\phi(0) = 0$ and $\phi(m) = n$.
 Considerable attention has been given to strategies for this minimization problem.

 In contrast, suppose we have represented the traces as parameterized plane curves in the basis $B_i(s)$, then we measure the squared distance as
 \begin{align}
     \langle x-u, x-u\rangle_B & + \langle y-v, y-v\rangle_B \notag \\
     &= \sum_{i=0}^d \sum_{j=0}^d (x_i - u_i)(x_j-u_j) \langle B_i, B_j\rangle_B
     + \sum_{i=0}^d \sum_{j=0}^d (y_i - v_i)(y_j-v_j) \langle B_i, B_j\rangle_B\notag \\
     &= \sum_{i=0}^d \left ( (x_i - u_i)^2  + (y_i - v_i)^2 \right ) \langle B_i, B_i\rangle_B.
 \end{align}
 This is straightforward and fast to compute since the $\langle B_i, B_i\rangle_B$ are a set of $d+1$ fixed values.
 
 \subsection{Legendre Inner Product and Legendre Basis}
 \label{sec:legendre-ip}
 The classical Legendre polynomials $P_n(x)$ are a degree-graded family orthogonal with respect to the following functional inner product
 \begin{equation}
     \langle f, g\rangle_P = \int_{-1}^1 f(x) g(x) dx
     \label{eq:legendre-ip}
 \end{equation}
 and normalized as
 \begin{equation}
     \langle P_m, P_n \rangle_P = 
     \begin{cases}
     \frac2{2n+1} &\text{if}~m = n \\
     0            &\text{otherwise}.
     \end{cases}
     \label{eq:legendre-normalization}
 \end{equation}
 We call \eqref{eq:legendre-ip} the \textit{Legendre inner product} and the corresponding norm is 
 \begin{equation}
     || f ||_P^2 = \langle f, f \rangle_P.
     \label{eq:legendre-norm}
 \end{equation}
 If $f$ and $g$ are polynomials given in Legendre basis such that
 \begin{align}
     f(x) &= \sum_{i=0}^d f_i P_i(x) & g(x) &= \sum_{i=0}^d g_i P_i(x),
     \label{eq:legendre-basis}
 \end{align}
 then, using \eqref{eq:legendre-normalization}, we have
 \begin{equation}
     \langle f, g\rangle_P = \sum_{i=0}^d \frac2{2i+1} f_i g_i.
     \label{eq:legendre-ip-formula}
 \end{equation}
 Note that the coefficients $f_i$ may be obtained as
 \begin{equation}
     f_i = \frac{\langle f, P_i\rangle_P}{\langle P_i,P_i\rangle_P}.
     \label{eq:legendre-coeffs}
 \end{equation}
 \subsection{Legendre-Sobolev Inner Product}
 \label{sec:legendre-sobolev-ip}
A {Sobolev inner product of order $N$} on a domain $D$ with Borel measures $\mu_i$ and coefficients $\lambda_i \in \mathbb{R}^+$ is defined as
\begin{equation*}
    \langle f, g \rangle_S = \sum_{i=0}^N \lambda_i \int_D f^{(i)} g^{(i)} d\mu_i.
\end{equation*}
Earlier work~\cite{golubitsky2010distance} examined classification using the \textit{Legendre-Sobolev} inner product studied by Althammer~\cite{Althammer},
\begin{equation}
    \langle f, g\rangle_{LS} = \int_{-1}^1 f(x) g(x) dx + \lambda \int_{-1}^1 f^\prime(x) g^\prime(x) dx.
    \label{eq:ls-ip}
\end{equation}

\section{Chebyshev-Sobolev Inner Product}
\label{sec:chebyshev-sobolev-ip}
We now come to the main topic of the article.
We propose to use a Sobolev inner product based on Chebyshev polynomials. Chebyshev polynomials are used extensively in numerical approximation due to their ``minimax'' property, that is minimizing the maximum error in the approximation of a function over an interval. As changing polynomial basis may be numerically ill conditioned, we explore using a Chebyshev basis \textit{ab initio} and working with this representation throughout.
\subsection{Chebyshev Inner Product}
We begin by noting the familiar properties of the Chebyshev polynomials of the first kind $T_n(x)$, which are a degree-graded family orthogonal with respect to the inner product
\begin{equation}
    \langle f, g\rangle_T = \int_{-1}^1 \frac{f(x) g(x)}{\sqrt{1-x^2}} dx
    \label{eq:chebyshev-ip}
\end{equation}
and normalized as
\begin{equation}
    \langle T_n, T_m\rangle_{T} = 
    \begin{cases}
    0          &\text{if~~} n \ne m \\
    \pi        &\text{if~~} n = m = 0 \\
    \frac \pi2 & \text{otherwise}.
\end{cases}
    \label{eq:chebyshev-normalization}
\end{equation}
We call~\eqref{eq:chebyshev-ip} the \textit{Chebyshev inner product} and the corresponding norm is
\begin{equation}
    ||f||_T^2 = \langle f, f \rangle_T.
    \label{eq:chebyshev-norm}.
\end{equation}
If $f(s)$ and $g(s)$ are polynomials of degree $d$ written in a Chebyshev basis as
\begin{align}
    f(x) &= \sum_{i=0}^d f_i T_i(x)  &
    g(x) &= \sum_{i=0}^d g_i T_i(x),
    \label{eq:chebyshev-basis}
\end{align}
then, using \eqref{eq:chebyshev-normalization}, we have
\begin{align}
    \langle f, g\rangle_T  &=
    \int_{-1}^1 \left (\sum_{i=0}^d \sum_{j=0}^d f_i T_i(x) g_j T_j(x)\right )\frac{dx}{\sqrt{1-x^2}}  \nonumber \\
    &= \int_{-1}^1 \left (\sum_{i=0}^d f_i g_i T_i(x) T_i(x) + \text{cross terms} \right )\frac{dx}{\sqrt{1-x^2}}  \nonumber \\
    &= \frac\pi2 \left ( 2 f_0 g_0 + \sum_{i=1}^d f_i g_i \right ).
    \label{eq:chebyshev-ip-formula}
\end{align}
The coefficients $f_i$ may be obtained as
 \begin{equation}
     f_i = \frac{\langle f, T_i\rangle_T}{\langle T_i,T_i\rangle_T}.
     \label{eq:chebyshev-coeffs}
 \end{equation}
\subsection{Chebyshev-Sobolev Inner Product}
We now consider the first order Sobolev inner product for Chebyshev polynomials,
\begin{align}
    \langle f,g \rangle_{CS} = \int_{-1}^{1}\frac {f(x)g(x)}{\sqrt{1-x^2}}ds + \lambda \int_{-1}^{1} \frac{f^\prime(x) g^\prime(x)}{\sqrt{1-x^2}}ds
\label{eq:chebyshev-sobolev-ip}
\end{align}
where $\lambda > 0$.   We call this a \textit{Chebyshev-Sobolev inner product} with corresponding norm
\begin{equation}
    || f ||_{CS}^2 = \langle f, f \rangle_{CS} .
\end{equation}
Applying Gram-Schmidt orthogonalization with respect to $\langle \cdot, \cdot\rangle_{CS}$ to monomials $\{1,x,x^2,....\}$ gives a dgree-graded family, we call \textit{Chebyshev-Sobolev polynomials}. Lets denote by $S_{\lambda\,i}$.  We take the same normalization as for Chebyshev polynomials so $S_{0~i} = T_i$.

An analytic function $f(x)$ may be written as a series using the basis $S_{\lambda\,i}$
\[
f(x) = \sum_{i=0}^\infty f_i S_{\lambda\,i}(s),
\]
where the coefficients $f_i$ satisfy
\[
f_i = \frac{\langle f, S_{\lambda\,i}\rangle_{CS}}
           {\langle S_{\lambda\,i}, S_{\lambda\,i}\rangle_{CS}}.
\]
Approximations to $f(x)$ may be obtained by summing only up to some degree $d$.

\section{A Matching Algorithm}
\label{sec:algorithm}

Comparing a sample input handwritten character to a set of model characters to find the best match can now be done as follows:
\begin{enumerate}
    \item Receive as inputs points $(x_i, y_i), i \in [0..N]$.
    \item Compute piece-wise polynomial splines (\textit{e.g.} piece-wise-linear or piece-wise-cubic) interpolating the sequences $x_i$ and $y_i$.
    \item Compute arc lengths $s_i, i \in[0..N]$ from the point $(x_0, y_0)$ to $(x_i, y_i)$ along the interpolating splines.   
    \item Shift and scale so that $s_0 = -1$ and $s_N = 1$.
    This scales the character to a standard size.
    \item Let $C_X(s)$ and $C_Y(s)$ be the piece-wise splines such that $C_X(s_i) = x_i$ and $C_Y(s_i) = y_i$.
    \item Compute the coefficients $X_i$ and $Y_i$, $i \in[1..d]$ such that
    $C_X(s) = \sum_{i=0}^d X_i T_i(s)$ and $C_Y(s) = \sum_{i=0}^d Y_i T_i(s)$.
    This requires $2N$  low-degree polynomial integrations done by formula.  The coefficients of $T_0$ are to be dropped to center the character.
    \item At this point, we have a resolution-independent representation of the character that is centered and scaled to a standard size.  This representation requires $2d$ numbers.
    \item Compare this sample character against the model characters one by one.   For each model character $G$ compute the euclidean norm as follows:
    \[
    D^2_G = || C_X - G_X ||^2 + || C_Y - G_Y ||^2
    \]
    and return the character model that gives the minimum.
\end{enumerate}

We remark on a few things:
\begin{itemize}
\item
All the approximation comes from choosing the truncation degree $d$ of the series.

\item
The choice of order of the interpolating spline should ideally be the same for the sample character as for all the models, but in practice it might not matter very much.
The step that scales $s$ is important.

\item
The integrations to obtain the coefficients in $C_X$ and $C_Y$ can be done by polynomial formulas and do not require numerical integration techniques.

\item If the summations to compute the Chebyshev basis coefficients of the derivative polynomials becomes computationally significant, then we can consider using Clenshaw summation~\cite{clenshaw}.

\item This whole procedure could use shifted Chebyshev polynomials $T^*_n(x) = T_n(2x-1)$ on $[0,1]$, in which case $s$ would be scaled to $[0,1]$.
\end{itemize}

The main point with this, and with the Legendre-Sobolev method, is that 
it is possible to compare efficiently against a great many models, since the computation for each depends only on the degree $d$ of the approximation and in particular does not depend on the number of sample points $N$.

\section{Experiments on Digital Ink}
\label{sec:experiments}
To give a first assessment of the suitability of the Chebyshev-Sobolev polynomials, we have run tests on Ink ML dataset \cite{InkML} and UCI pendigits dataset \cite{misc_pen-based_recognition_of_handwritten_digits_81}. The ink dataset consist of traces of different letters and UCI pendigits dataset contain samples of handwritten digits (0-9) of multiple users. We started the experiment by approximating the digital ink using Chebyshev-Sobolev polynomials. 
For the sake of comparison, we have also considered Legendre, Chebyshev, and Legendre-Sobolev polynomial. We perform the following experiments:
\subsection{Handwriting representation using Chebyshev-Sobolev series}
\begin{figure}[htbp]
    \centering\
    \begin{subfigure}[b]{0.45\textwidth}
        \centering
        \includegraphics[width=\textwidth, height = 3cm]{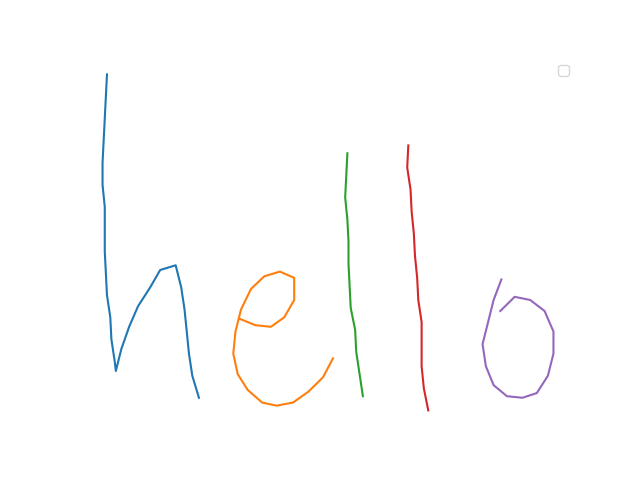}
        \caption{Orignal}
        \label{hello_plot}
    \end{subfigure}
    \hfill
    \begin{subfigure}[b]{0.45\textwidth}
        \centering
        \includegraphics[width=\textwidth, height = 3cm]{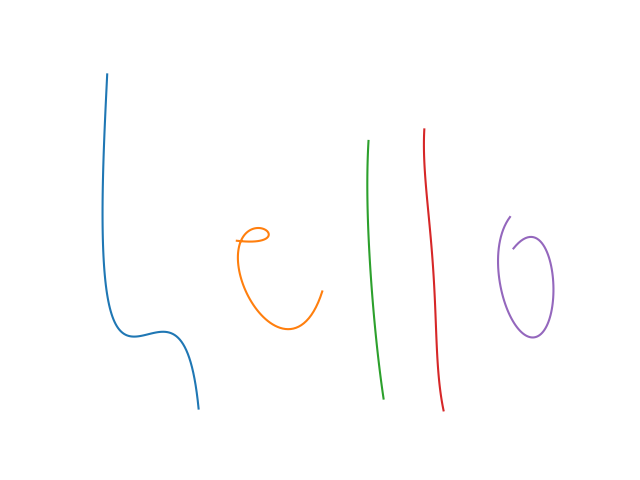}
        \caption{Approximated with degree = 5}
        \label{hello+plot_approx}
    \end{subfigure}

    \begin{subfigure}[b]{0.45\textwidth}
        \centering
        \includegraphics[width=\textwidth, height = 3cm]{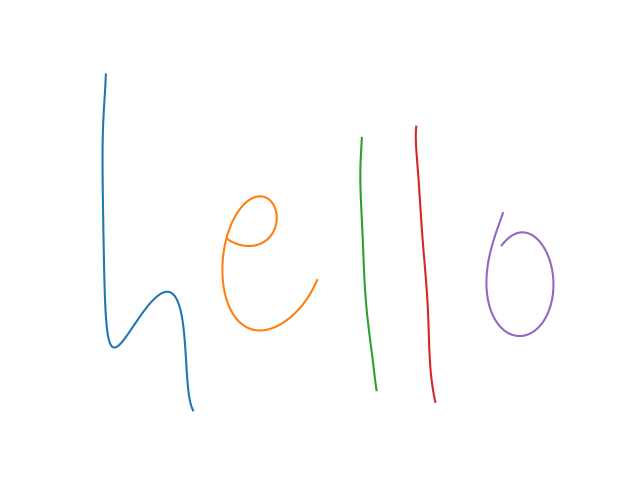}
        \caption{Approximated with degree = 7}
        \label{hello+plot_approx_1}
    \end{subfigure}
    \hfill
    \begin{subfigure}[b]{0.45\textwidth}
        \centering
        \includegraphics[width=\textwidth, height = 3cm]{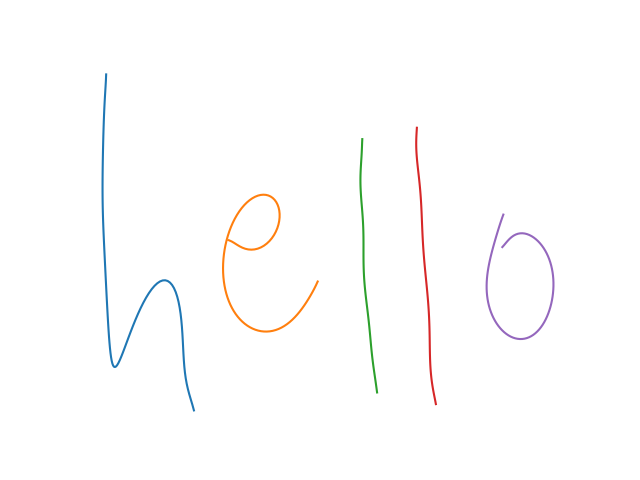}
        \caption{Approximated with degree = 10}
        \label{hello+plot_approx_2}
    \end{subfigure}
    
    \caption{Handwriting representation test on Ink ML dataset}
    \label{fig:comparison_one}
\end{figure}

In this experiment, we plot the approximated curve considering different degrees for the orthogonal polynomials. We have considered the hyper parameter $\lambda$ as $\frac{1}{8}$
 and compare the approximated curve with the original trace by increasing degree. It can be observed that the quality of approximated curve improves with increase in degree.
Figure \ref{hello_plot} shows the letters in original trace of Ink ML dataset. In this figure, we can observe large number of points resulting in shorter distance between the adjacent curves. From figure \ref{hello+plot_approx},\ref{hello+plot_approx_1}, and \ref{hello+plot_approx_2} it can be seen that the approximated curve are able to accurately and smoothly represent the letters. Additionally, with Chebyshev-Sobolev polynomials, we are able to accurately represent the self overlapping letters like "e", and "o".

\begin{figure}[htbp]
  \centering
  \begin{subfigure}[b]{\textwidth}
    \centering
    \includegraphics[width=\textwidth, height = 1.5 cm]{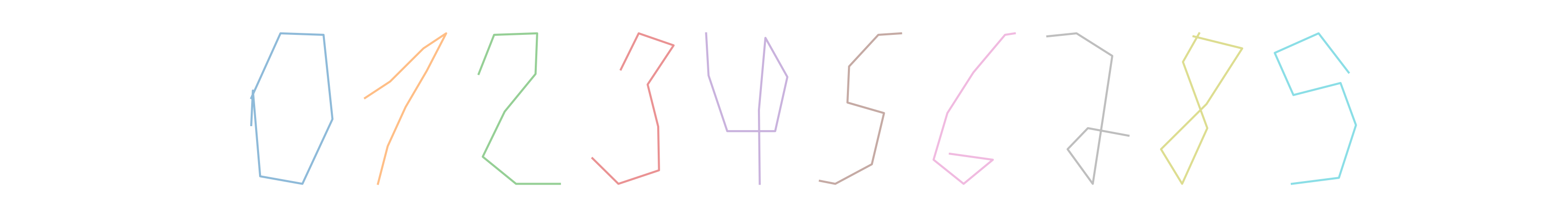} 
    \caption{Original }
    \label{UCI_dataset_orig}
  \end{subfigure}
  
  \vspace{0.5 cm} 
  
  \begin{subfigure}[b]{\textwidth}
    \centering
    \includegraphics[width=\textwidth, height = 1.5 cm]{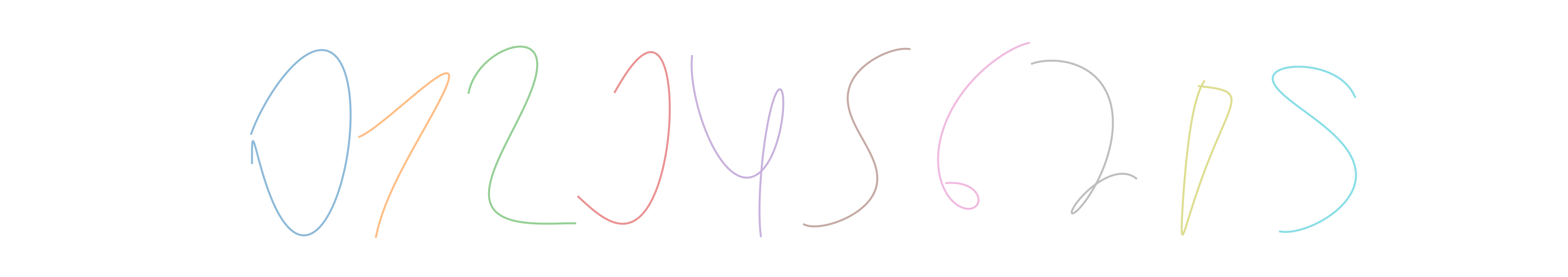} 
    \caption{Approximated digits with degree = 5}
    \label{UCI_dataset_approx_5}
  \end{subfigure}

\vspace{0.5cm} 
  
  \begin{subfigure}[b]{\textwidth}
    \centering
    \includegraphics[width=\textwidth, height = 1.5 cm]{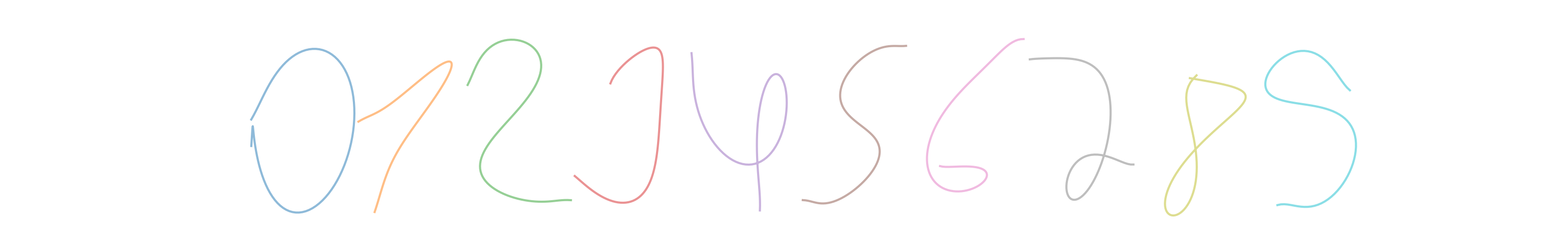} 
    \caption{Approximated digits with degree = 7}
    \label{UCI_dataset_approx_7}
  \end{subfigure}

  \vspace{0.5cm} 
  
  \begin{subfigure}[b]{\textwidth}
    \centering
    \includegraphics[width=\textwidth, height = 1.5 cm]{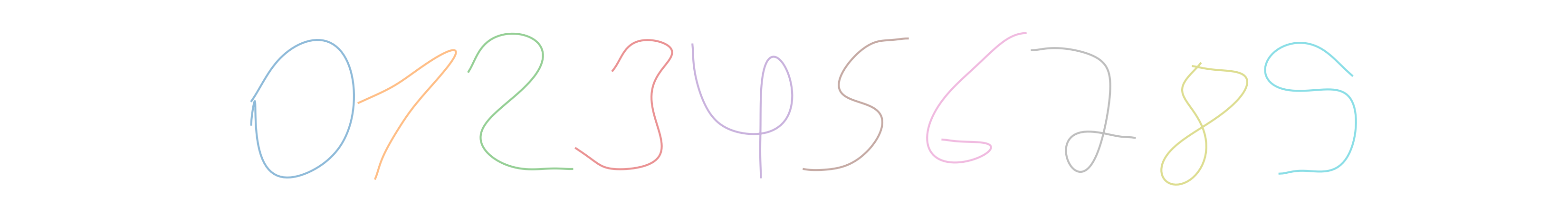} 
    \caption{Approximated digits with degree = 10}
    \label{UCI_dataset_approx_10}
  \end{subfigure}
\caption{Handwriting representation test on UCI pen digits dataset}
  \label{fig:digit_comparison_one}
\end{figure}

Figure \ref{UCI_dataset_orig} shows the 10 digits selected at random from the UCI pen digits dataset. In this figure we can observe that each digit has exactly 8 points and we connected these points considering them as linearly separable. Due to limited number of points and equal distances in the points, the handwritten digits are coarsely represented. From figure \ref{UCI_dataset_approx_5}, \ref{UCI_dataset_approx_7}  and \ref{UCI_dataset_approx_10}, it can be seen that the approximated curve are able to accurately represent the handwritten digits. The gaps between the point is smooth and there are no vibrations near the end points. 

\subsection{Representation error of approximated curve using Chebyshev-Sobolev series}
\begin{figure}[htbp]
    \begin{subfigure}[b]{\textwidth}
        \centering
        \includegraphics[width=\textwidth, height = 1.5 cm]{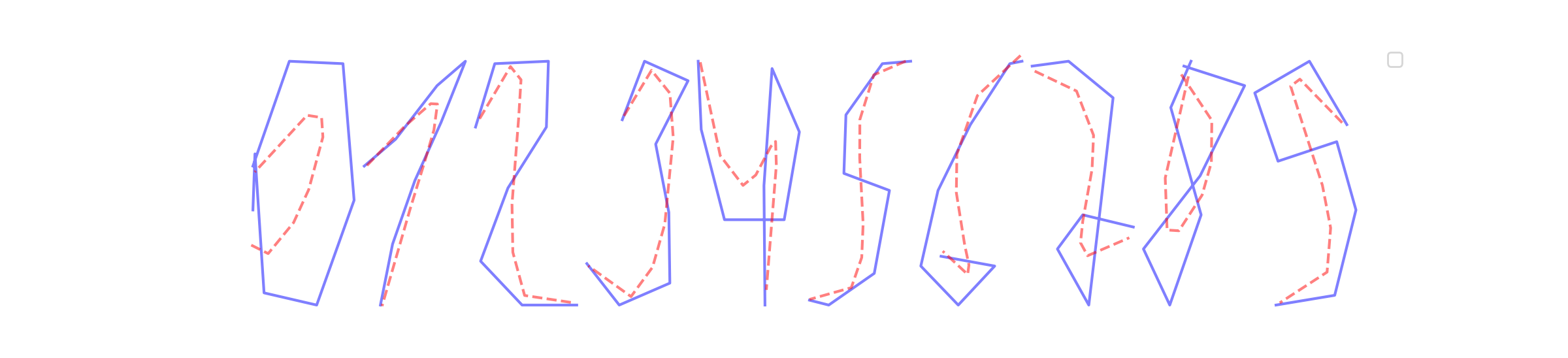} 
        \caption{Approximated digits with degree = 3}
        \label{UCI_dataset_approx_3_points}
    \end{subfigure}

    \vspace{1cm} 

    \begin{subfigure}[b]{\textwidth }
        \centering
        \includegraphics[width=\textwidth, height = 1.5 cm]{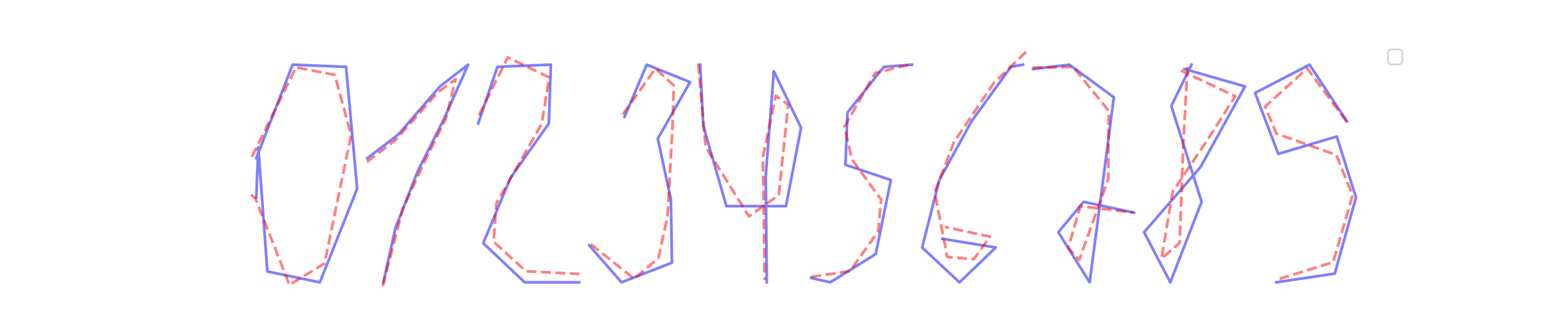} 
        \caption{Approximated digits with degree = 7}
        \label{UCI_dataset_approx_7_points}
    \end{subfigure}

    \vspace{1cm} 

    \begin{subfigure}[b]{\textwidth}
        \centering
        \includegraphics[width=\textwidth, height = 1.5 cm]{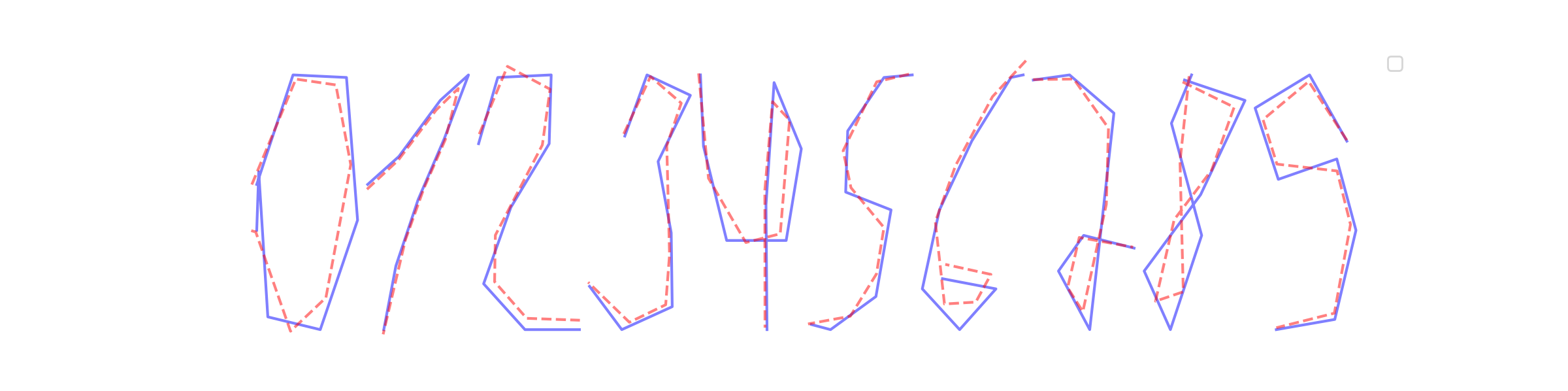} 
        \caption{Approximated digits with degree = 10}
        \label{UCI_dataset_approx_10_points}
    \end{subfigure}

    \vspace{1cm} 

    \begin{subfigure}[b]{\textwidth}
        \centering
        \includegraphics[width=\textwidth, height=1.5 cm]{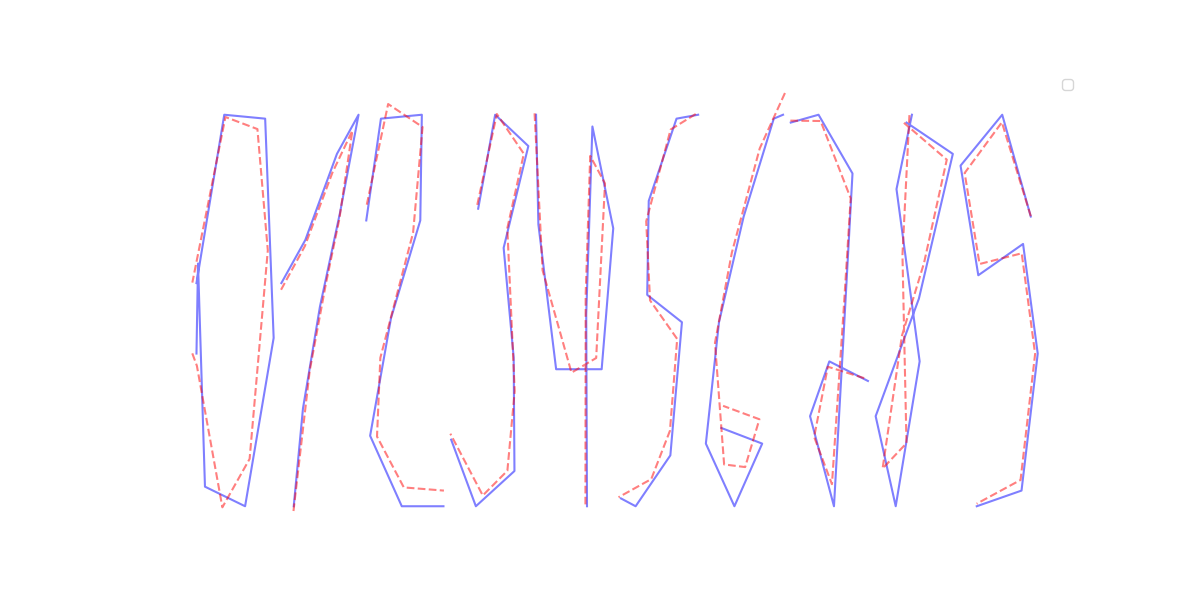} 
        \caption{Approximated digits with degree = 15}
        \label{UCI_dataset_approx_15_points}
    \end{subfigure}

    \caption{Representation error on UCI pen digits dataset}
    \label{fig:digit_comparison_two}
\end{figure}

\begin{figure}[htbp]
    \centering\
    \begin{subfigure}[b]{0.45\textwidth}
        \centering
        \includegraphics[width=\textwidth, height = 2 cm]{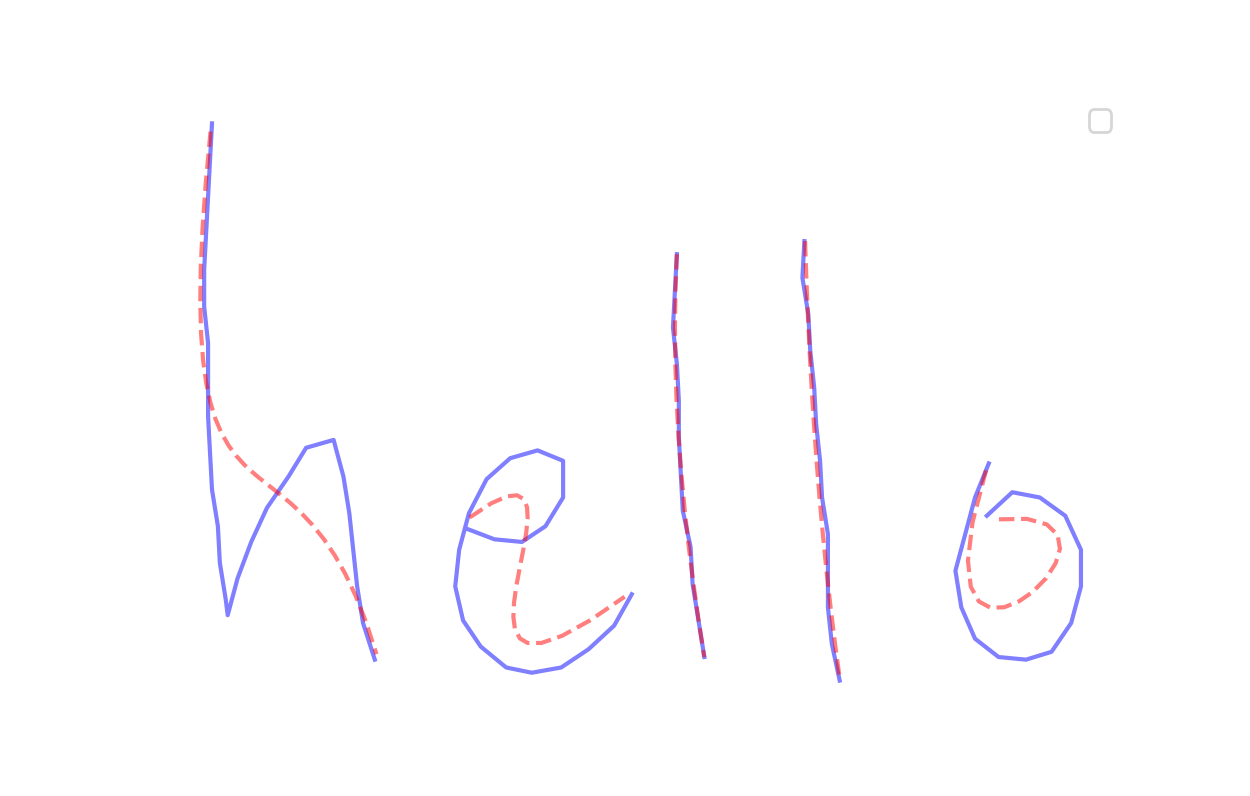}
        \caption{Approximated with degree = 3}
        \label{hello_3}
    \end{subfigure}
    \hfill
    \begin{subfigure}[b]{0.45\textwidth}
        \centering
        \includegraphics[width=\textwidth, height = 2 cm]{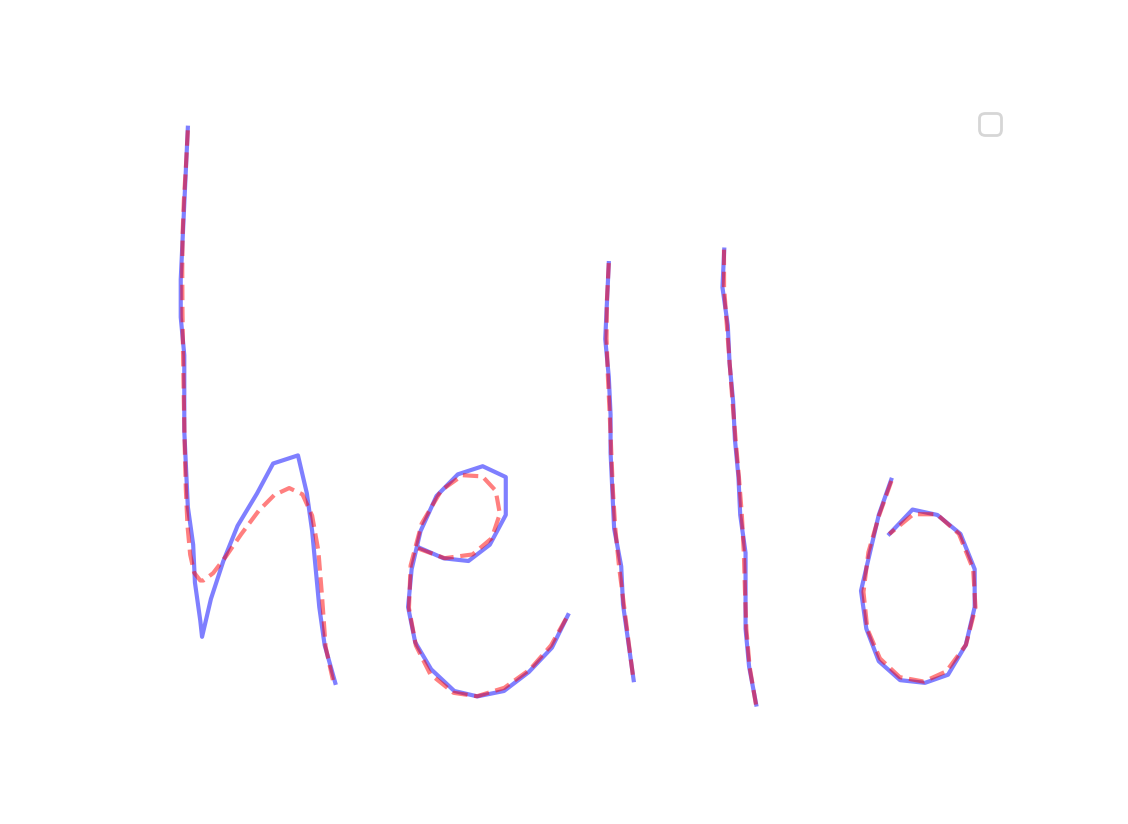}
        \caption{Approximated with degree = 7}
        \label{hello_7}
    \end{subfigure}
    \hfill
    \begin{subfigure}[b]{0.45\textwidth}
        \centering
        \includegraphics[width=\textwidth, height = 2 cm]{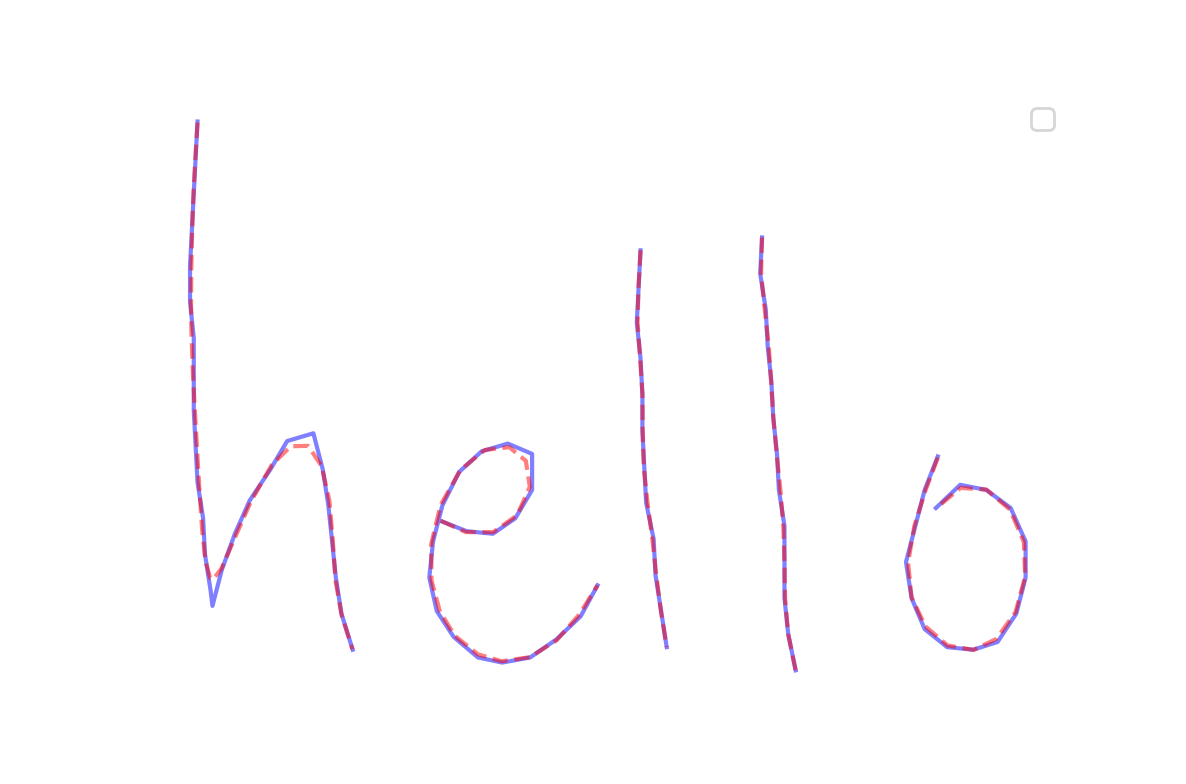}
        \caption{Approximated with degree = 10}
        \label{hello_10}
    \end{subfigure}
    \hfill
    \begin{subfigure}[b]{0.45\textwidth}
        \centering
        \includegraphics[width=\textwidth, height = 2 cm]{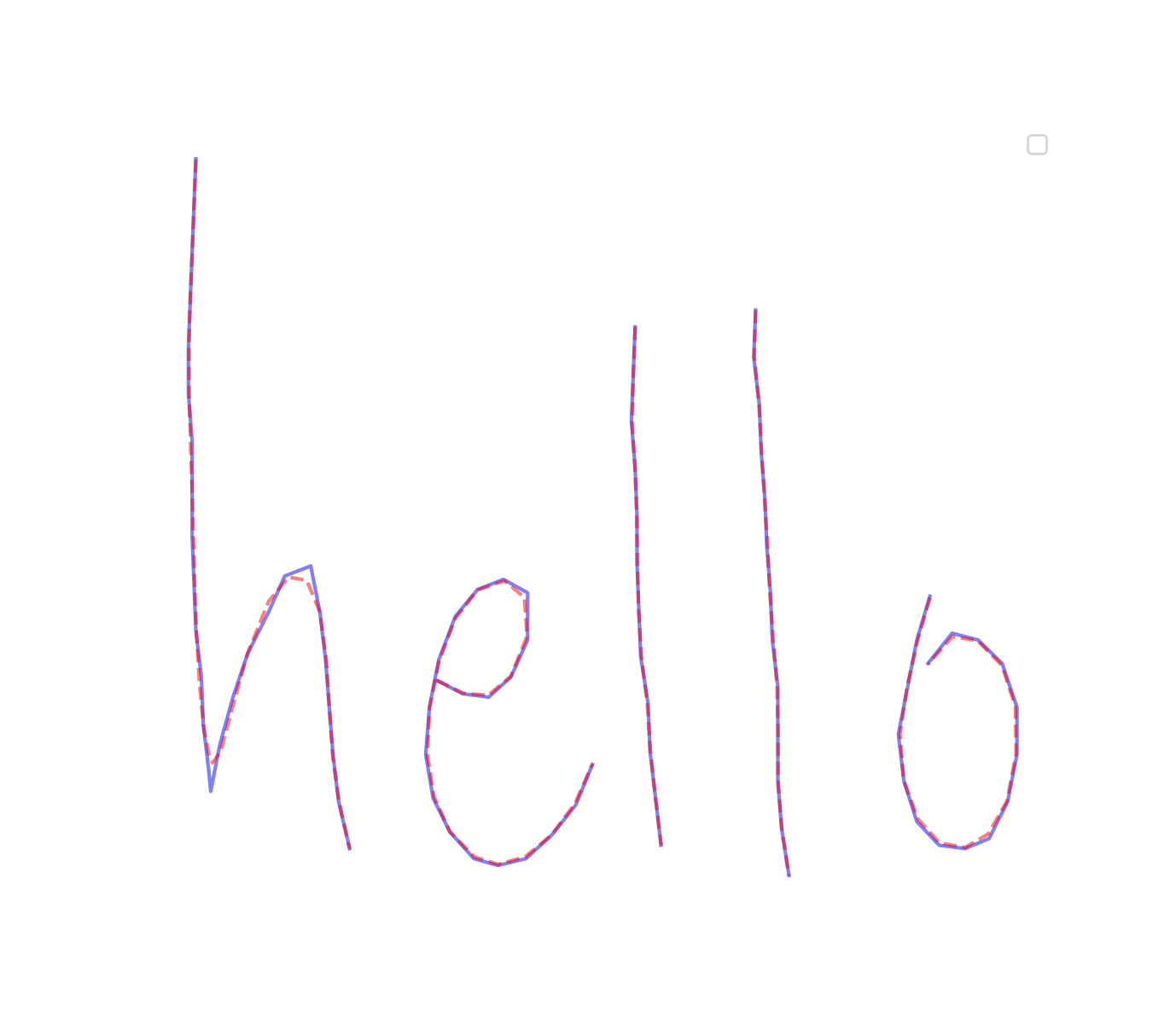}
        \caption{Approximated with degree = 15}
        \label{hello_15}
    \end{subfigure}
    \caption{Representation error on Ink ML dataset}
    \label{fig:comparison_two}
\end{figure}

\begin{figure}[htbp]
    \centering\
    \begin{subfigure}[b]{0.65\textwidth}
        \centering
        \includegraphics[width=9cm]{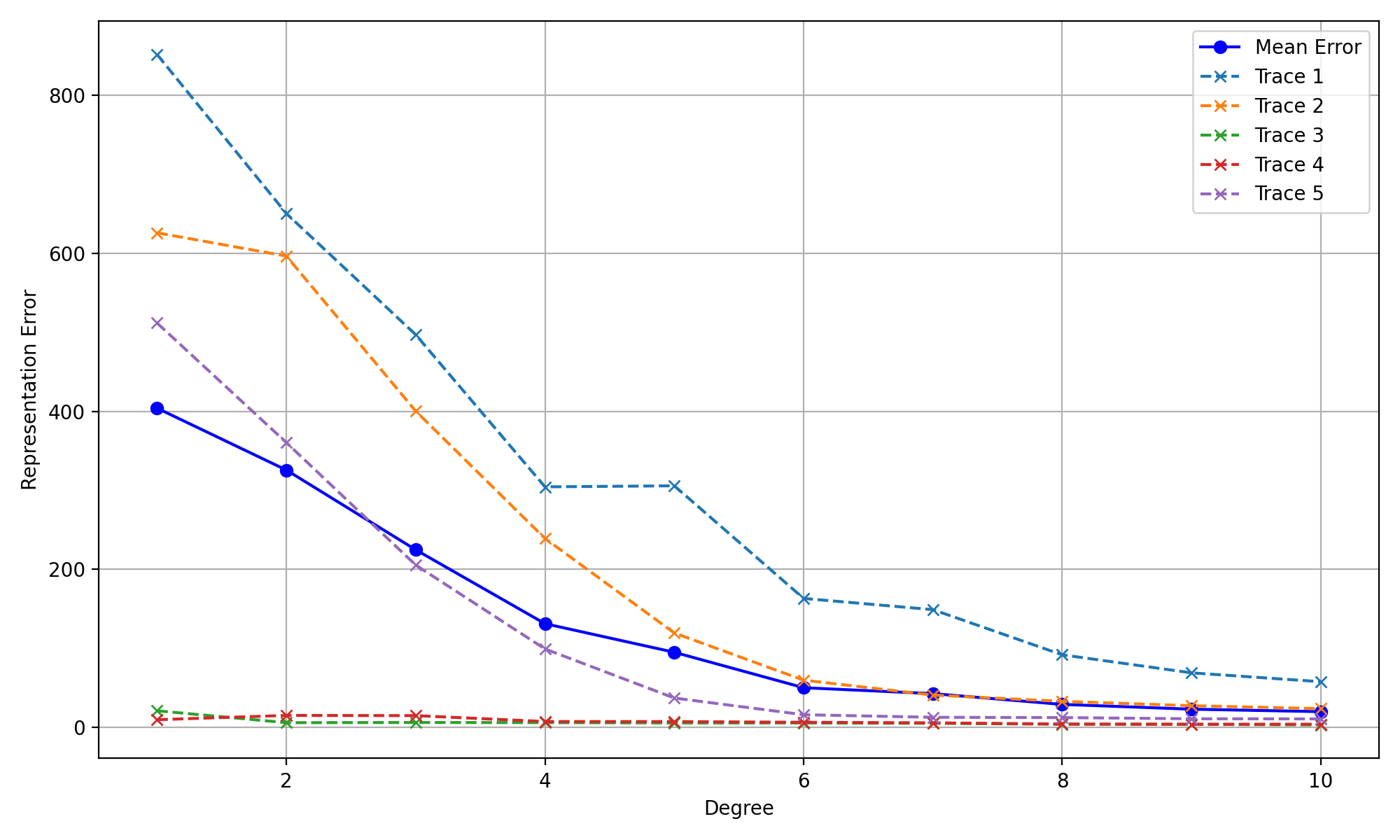}
        \vspace{-0.8\baselineskip}
        \caption{Representation error vs degree on InkML dataset}
        \label{repr_err_ink}
    \end{subfigure}
    
    \begin{subfigure}[b]{0.65\textwidth}
        \centering
        \includegraphics[width=8.5cm]{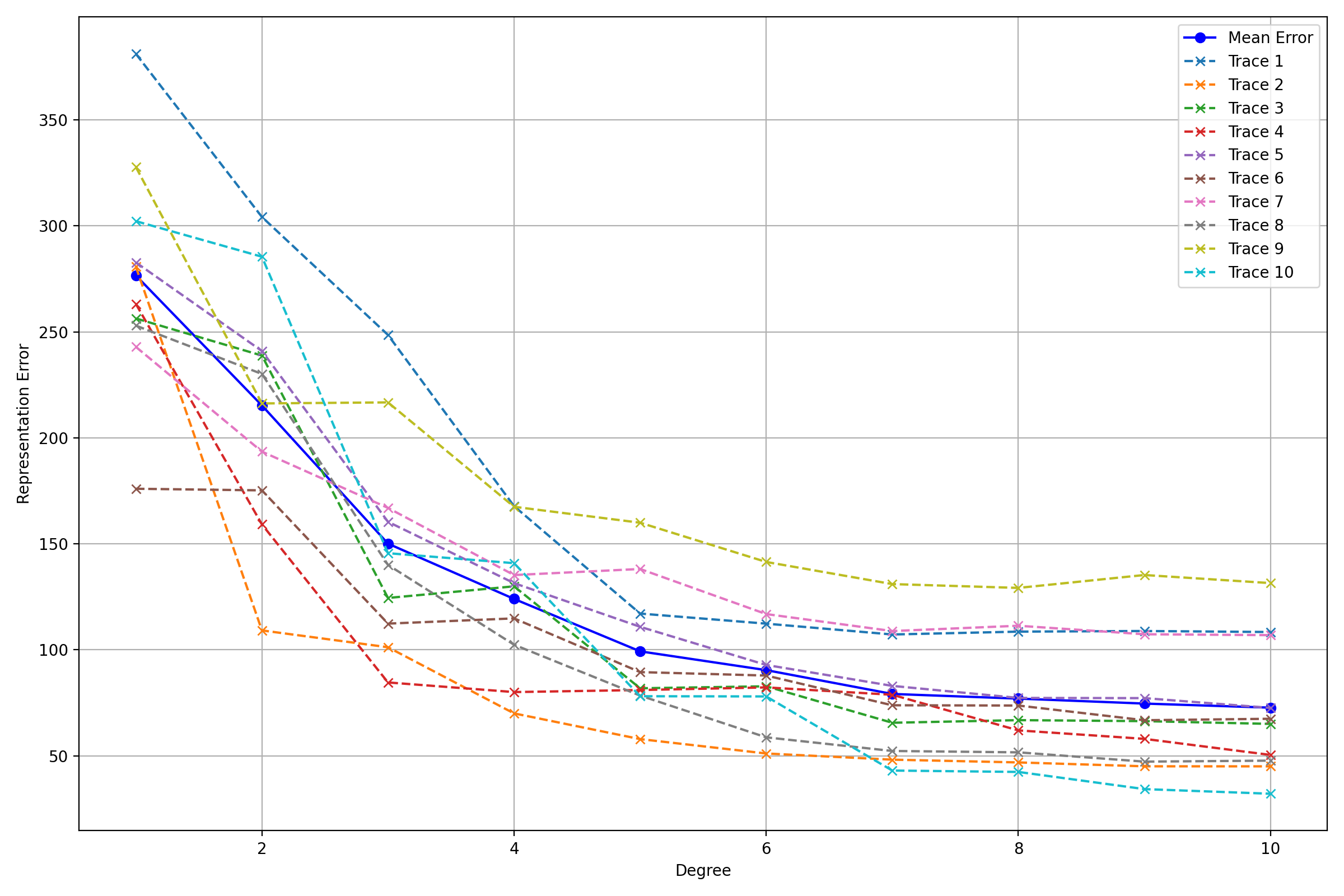}
        \vspace{-0.8\baselineskip}
        \caption{Representation error vs degree on UCI pendigits dataset}
        \label{repr_err_uci}
    \end{subfigure}
    \caption{Representation error using Chebyshev-Sobolev series}
\end{figure}

In this test, we measure the representation error by calculating euclidean distance from the original points and approximated points. The error can be represented as:
\begin{align}\label{err_repr}
    Error := \sum_{i=0}^N\sqrt{(x_i -\hat{x}_i)^2 + (y_i -\hat{y}_i)^2 }
\end{align}
where, N is the total number of points in the original trace, $(x_i,y_i)$ are the original points, and $(\hat{x}_i,\hat{y}_i)$ are the approximated points in the trace.
Figure \ref{UCI_dataset_approx_3_points}, \ref{UCI_dataset_approx_7_points}, \ref{UCI_dataset_approx_10_points}, and \ref{UCI_dataset_approx_15_points} represents the points generated by considering Chebyshev-Sobolev polynomials upto degree 3, 7, 10, and 15 respectively. For the sake of comparison we have joined the points with the straight line to see likelihood with the original trace. The approximated points for some traces like of "1" are very close with degree 7, while some self looping traces like "9" need higher degree polynomials for approximation. These findings are also justified in Figure \ref{repr_err_uci} which is plot of representation error in equation (\ref{err_repr}) vs degree of polyomial.
Similarly, the Figure \ref{hello_3}, \ref{hello_7}, \ref{hello_10}, and \ref{hello_15} shows approximated points of Ink ML dataset considering polynomials upto degree 3, 7, 10, and 20. Additionally, figure \ref{repr_err_ink} and \ref{repr_err_uci} shows the plot of representation error vs degree.

\enlargethispage{4\baselineskip}
\subsection{Handwriting recognition using $k$ nearest neighbours}
\begin{figure}[htbp]
    \centering
    \begin{subfigure}[b]{0.65\textwidth}
        \centering
        \includegraphics[width=11cm]{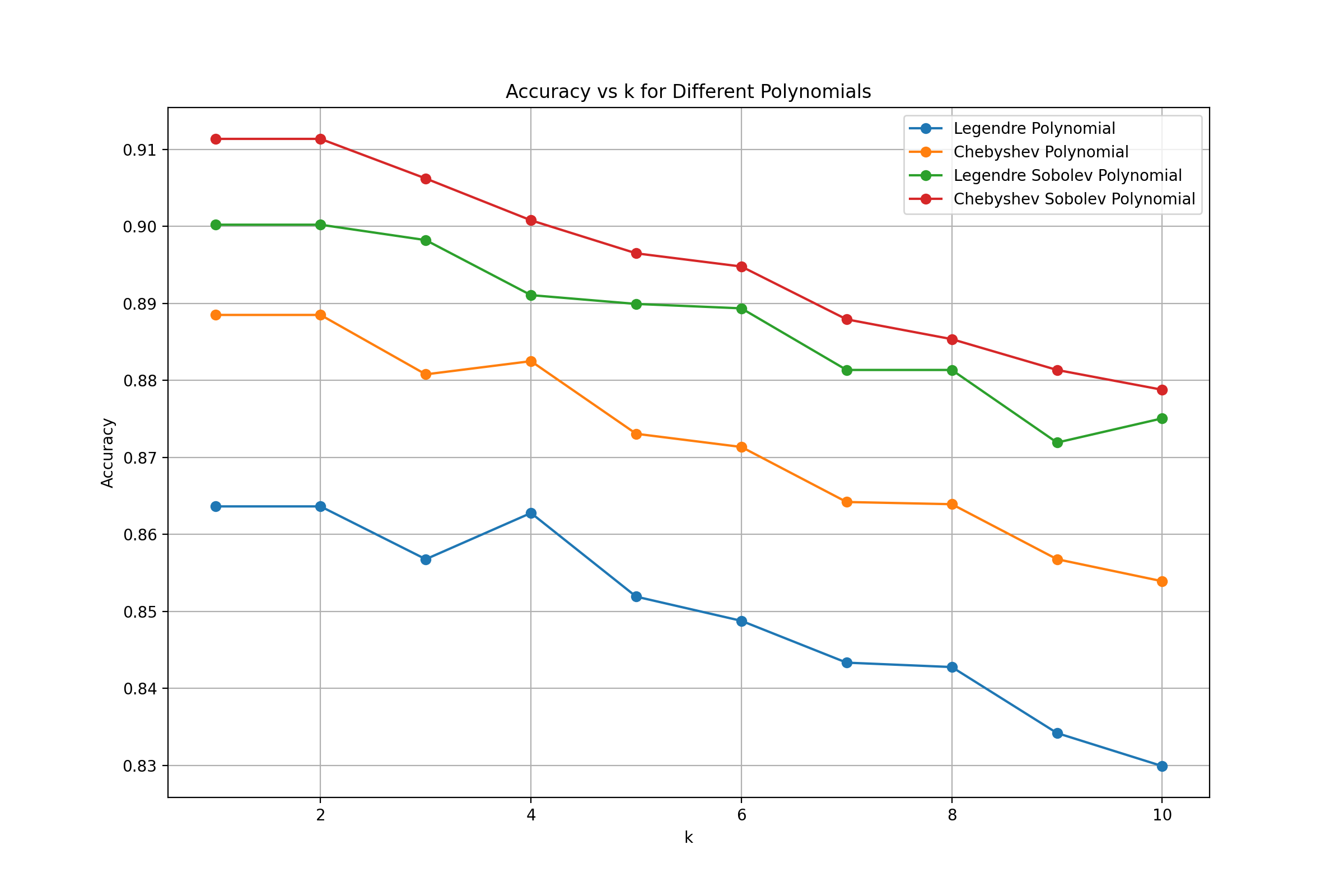}
        \vspace{-1.7\baselineskip}
        \caption{Accuracy vs $k$ for Different Polynomials}
        \label{k_accuracy}
    \end{subfigure}
    
    \begin{subfigure}[b]{0.65\textwidth}
        \centering
        ~\hspace{.7cm}\includegraphics[width=9.3cm]{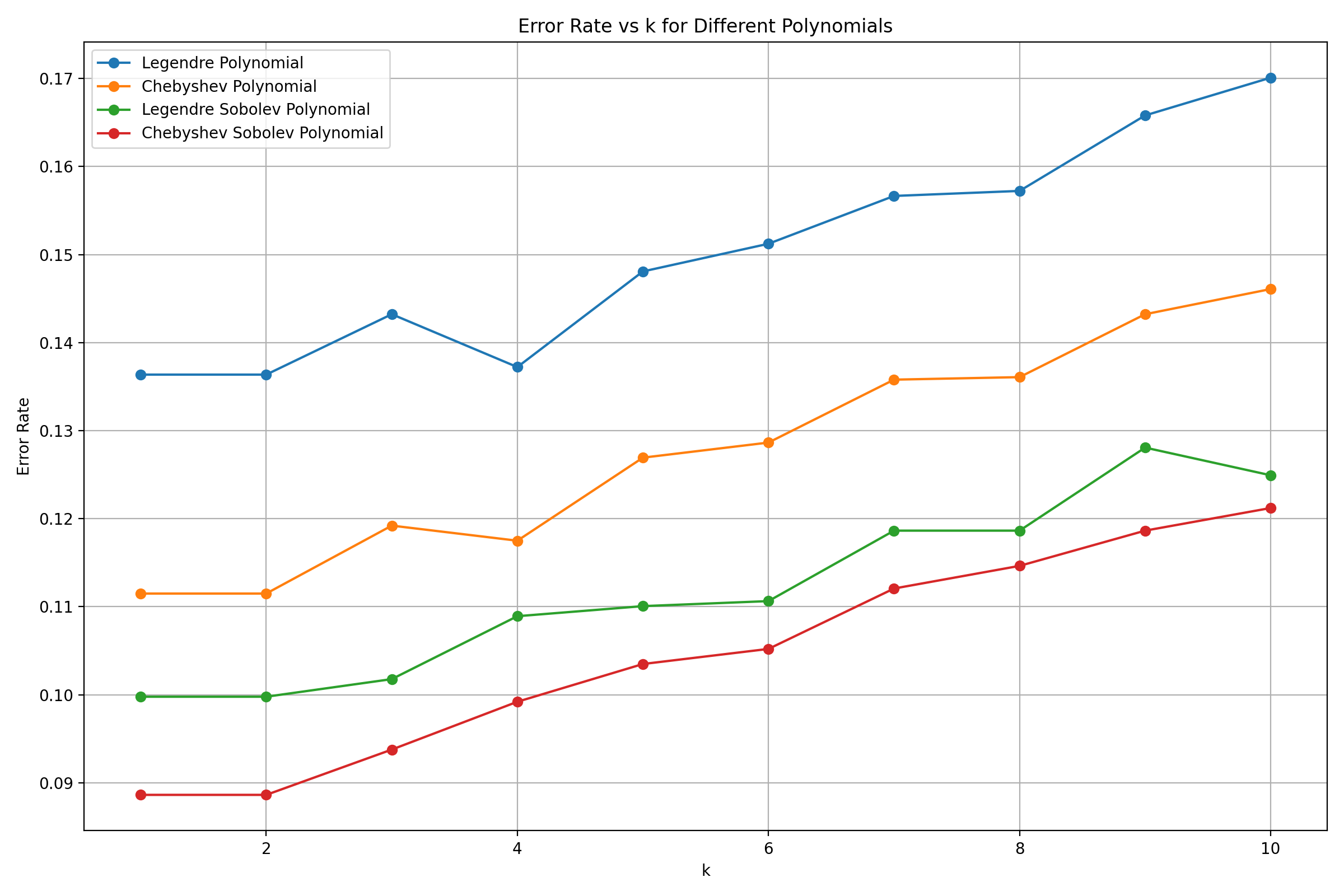}
        \vspace{-1.7\baselineskip}
        \caption{Error Rate vs $k$ for Different Polynomials}
        \label{error_k}
    \end{subfigure}
    \caption{Handwriting recognition using $k$ nearest neighbours}
\end{figure}

In order to evaluate the effectiveness of handwriting recognition using Chebyshev-Sobolev series we perform a test using UCI pendigits dataset from \cite{misc_pen-based_recognition_of_handwritten_digits_81}. The dataset consist of 10992 samples of 10 different classes of numerical digits (0-9). Each digit consist of equal 8 points and each class has approximately 1000 samples. 
In order to examine the classification performance of Chebyshev-Sobolev series, we used two third of the handwritten digits in the data set
for training, the remaining ones for testing, and then computed the
coefficients of these handwritten digits respectively. We have truncated the series upto $d$=10 and used $\lambda=\frac{1}{8}$. 
Afterwards, Euclidean distance was used as the metric to perform
k-nearest neighbours classification of the test set from the training set,
where $k$ ranges from 1 to 10. Fig. \ref{k_accuracy} and Fig. \ref{error_k} shows the accuracy and error rate for $k$ ranging from 1 to 10 for Legendre , Chebyshev, Legendre- Sobolev, and Chebyshev-Sobolev Polynomial. The accuracy remains high, but decreases with increase in $k$ for all the methods. The error rates increases with increase in $k$. This is an expected behaviour of $k$-NN method where the smaller $k$ captures the best trends of the data.
It can be Observed that Chebyshev-Sobolev polynomials performs the best for all the values of $k$.
Also, Legendre-Sobolev has the accuracy higher than Chebhyshev Polynomial and Legendre Polynomials.
The decrease in accuracy with $k$ is comparatively gradual for Chebyshev-Sobolev polynomial.

\section{Conclusions}
\label{sec:conclusions}
We have defined the family of Chebyshev-Sobolev polynomials in analogy to the Legendre-Sobolev polynomials studied by Althammer and used in earlier work in mathematical handwriting recognition.   Some initial results show that they can be superior to Legendre-Sobolev polynomials in some circumstances.  This motivates us to study their properties in more detail and their effectiveness on more varied symbol corpora.



\clearpage
\IfFileExists{IfExistsUseBBL.bbl}{%

}{%
\bibliography{main}
}


\end{document}